\documentclass[10pt,twocolumn,letterpaper]{article}

\usepackage{wacv}              

\usepackage[accsupp]{axessibility}

\usepackage{graphicx}
\usepackage{amsmath}
\usepackage{amssymb}
\usepackage{booktabs}
\usepackage{adjustbox}
\usepackage{multirow}
\usepackage{arydshln}
\usepackage{xcolor}
\usepackage{float}

\usepackage[pagebackref,breaklinks,colorlinks]{hyperref}

\usepackage[capitalize]{cleveref}
\crefname{section}{Sec.}{Secs.}
\Crefname{section}{Section}{Sections}
\Crefname{table}{Table}{Tables}
\crefname{table}{Tab.}{Tabs.}

\def\wacvPaperID{1736} 
\def\confName{WACV}
\def\confYear{2024}

\begin{document}

\title{Learning Generalizable Perceptual Representations for Data-Efficient No-Reference Image Quality Assessment}

\author{Suhas Srinath$^*$\quad Shankhanil Mitra$^*$\quad Shika Rao \quad Rajiv Soundararajan\\
Indian Institute of Science\\
Bengaluru, India 560012\\
{\tt\small \{suhass12, shankhanilm, shikarao, rajivs\}@iisc.ac.in}
}
\maketitle
\def\thefootnote{*}\footnotetext{Equal contribution.}
\def\thefootnote{$\S$}\footnotetext{\url{https://github.com/suhas-srinath/GRepQ}}
\begin{abstract}
    No-reference (NR) image quality assessment (IQA) is an important tool in enhancing the user experience in diverse visual applications. A major drawback of state-of-the-art NR-IQA techniques is their reliance on a large number of human annotations to train models for a target IQA application.
    To mitigate this requirement, there is a need for unsupervised learning of generalizable quality representations that capture diverse distortions. We enable the learning of low-level quality features agnostic to distortion types by introducing a novel quality-aware contrastive loss. Further, we leverage the generalizability of vision-language models by fine-tuning one such model to extract high-level image quality information through relevant text prompts. The two sets of features are combined to effectively predict quality by training a simple regressor with very few samples on a target dataset. Additionally, we design zero-shot quality predictions from both pathways in a completely blind setting. 
    Our experiments on diverse datasets encompassing various distortions show the generalizability of the features and their superior performance in the data-efficient and zero-shot settings. 
\end{abstract}

\section{Introduction}\label{sec:introduction}

The increasing number of imaging devices, including cameras and smartphones, has significantly increased the volume of images captured, edited, and shared on a global scale. 
As a result, there is a necessity to assess the quality of visual content to enhance user experience.
Image Quality Assessment (IQA) is generally divided into two categories: full-reference (FR) and no-reference (NR) IQA. 
While FR-IQA relies on pristine reference images for quality assessment, NR-IQA is more relevant and challenging due to the absence of clean references for user-captured images. 
 
Most successful NR IQA methods are deep-learning based, and require a large number of images with human opinion scores for training. As imaging systems evolve, the distortions also evolve, making it difficult to keep creating large annotated datasets for training NR IQA models. This motivates the study of limited data or data-efficient NR IQA models which can be trained on a target IQA application (or database) with limited labels. Such an approach works best if the learned quality representations can generalize well across different distortion types for various IQA tasks. These representations can then be mapped to quality using a simple linear model \cite{contrique, reiqa} using the limited labels on a target application. Further, it is desirable that we learn these representations without requiring any human annotations of image quality. The goal of our work is to learn generalizable image quality representations through self-supervised learning to design data-efficient NR models for a target IQA application.  

In this regard, while DEIQT\cite{deiqt} studies the data-efficient IQA problem, they train the entire network with millions of parameters, which still requires a reasonable number of labeled training images. On the other hand, recent methods such as CONTRIQUE \cite{contrique}, Re-IQA \cite{reiqa}, and QPT \cite{qap} focus on self-supervised contrastive learning to learn quality features, which can potentially yield superior performance with limited labels. 
However, these methods do not consider that images with different distortions could have the same quality, thereby limiting the generalizability of their image features across varied distortions. 

Our main contribution is the design of Generalizable Representations for Quality (\textbf{GRepQ}) that can predict quality by training a simple linear model with few annotations. 
We present two sets of features, one to capture the local low-level quality variations and another to predict quality using the global context. 
To capture low-level quality features, we propose a quality-aware contrastive learning strategy guided by a perceptual similarity measure between distorted versions of an image.
In particular, we bring similar-quality images closer in the latent space irrespective of their distortion types. 
This is achieved by assigning a weight based on a similarity measure between every pair of distorted versions of an image. 
Our strategy enables the learning of generalizable quality representations invariant to distortion types.

We also leverage the generalization capabilities of large vision-language models for extracting high-level quality information. Notably, the versatile CLIP\cite{clip} model can be applied to zero-shot quality prediction \cite{clipiqa}, although a lack of task-specific fine-tuning limits it. 
While LIQE \cite{liqe} fine-tunes CLIP by integrating scene and distortion information, it requires large-scale training with human labels. 
We overcome these limitations through a novel unsupervised fine-tuning of CLIP. 
We achieve this by segregating images of higher and lower quality into groups using antonym text prompts and employing a group-contrastive loss with respect to the prompts. 
Our group-contrastive learning facilitates the learning of high-level quality representations that can generalize well to diverse content and distortions.

The features from both pathways can be combined to learn a simple regressor trained with few samples from any IQA dataset. Additionally, predictions can be made in a zero-shot setting using the learned features, which can then be combined to provide a single objective score. We show 
through extensive experiments that our framework shows superior performance in both the data-efficient as well as zero-shot settings. We summarize the main contributions of our framework as follows:

\begin{itemize}
    \item A quality-aware contrastive loss that weighs positive and negative training pairs using a ``soft" perceptual similarity measure between a pair of samples to enable representation learning invariant to distortion types.  
    \item An unsupervised task-specific adaptation of a vision-language model to capture semantic quality information. We achieve this by separating higher and lower-quality groups of images based on quality-relevant antonym text prompts.  
    \item Superior performance of our method over other NR-IQA methods trained using few samples (data-efficient) on several IQA datasets to highlight the generalizability of our features. Additionally, we show superior cross-database prediction performance. 
    \item A zero-shot quality prediction method using the learned features and its superior performance compared to other zero-shot (or completely blind) methods. 
\end{itemize}

\section{Related Work}\label{sec:related_work}

\subsection{Supervised NR-IQA}\label{sec:rel_nriqa}

Many popular supervised NR-IQA methods such as BRISQUE\cite{brisque}, DIIVINE\cite{diivine}, BLIINDS\cite{bliinds}, CORNIA\cite{cornia} predict quality using hand-crafted natural scene statistics based features. Such methods have succeeded when images contain synthetic distortions but often suffer when the distortions are more complex or authentic. To mitigate this, several deep learning-based methods have emerged that are either trained in an end-to-end fashion \cite{dbcnn, e2e2, e2e3} or use a pre-trained feature encoder that can be fine-tuned for IQA \cite{dbcnn}. Further, transformer-based models have shown promise on authentic and synthetically distorted images \cite{transformers1, transformers2, transformers3, hyperiqa}. Methods such as MetaIQA\cite{metaiqa} employ meta-learning to learn from synthetic data and adapt to real-world images efficiently. A recent method, LIQE \cite{liqe} adapts the CLIP model for IQA via scene and distortion classification along with supervised fine-tuning on several IQA datasets. 
However, the model requires multiple annotations per image during training making the model infeasible when adapting to newer and more complex datasets in the data-efficient regime.

\subsection{Self-Supervised Quality Feature Learning}
Although supervised NR-IQA methods have shown reasonable performance in quality prediction, they still possess the limitation of requiring large amounts of human annotations for training. One of the earliest approaches in this domain was through the design of quality-aware codebooks \cite{cornia}. Later, different ranking-based methods were used for quality-aware pre-training \cite{rankiqa}. Contrastive learning-based training such as CONTRIQUE \cite{contrique}, Re-IQA \cite{reiqa} and QPT \cite{qap} learn quality representations by contrasting multiple levels of synthetic distortions. While Re-IQA \cite{reiqa} also uses high and low-level features, our method significantly differs from Re-IQA in how the low-level and high-level features are designed.  Further, all the above methods neither consider the generalizability to unseen distortions nor do they consider the data-efficient evaluation setting.

\subsection{Zero-Shot or Completely Blind (CB) IQA}\label{sec:rel_cbiqa}

Another class of IQA methods are zero-shot or completely blind and do not require any human opinions for their design. For example, NIQE \cite{niqe} neither requires training on a dataset of annotated images nor knowledge about possible degradations.  
IL-NIQE \cite{ilniqe} improves over NIQE by integrating other quality-aware features based on Gabor filter responses, gradients, and color statistics.
However, both methods tend to fail on authentic and other complex distortions.
A recent method \cite{clmi} learns deep features using contrastive learning to predict quality without any supervision. 
However, the performance shown on in-the-wild IQA datasets still provides scope for further improvement.
Leveraging the contextual information from CLIP\cite{clip}, CLIP-IQA \cite{clipiqa} shows that a zero-shot application of the CLIP model can yield promising quality predictions. However, zero-shot methods tend to have poorer performance and motivate the use of limited labels on target IQA applications to improve performance. 

\subsection{Data-Efficient IQA}\label{sec:rel_data_eff}
IQA in the low-data setting remains relatively unexplored. 
Data-efficient image quality assessment (DEIQT) \cite{deiqt} shows that IQA models can be efficiently fine-tuned with very few annotated samples from a target dataset, enabling generalization through data efficiency.
Further, with a sufficient number of training samples, data-efficient training can achieve performances of full dataset supervision on multiple IQA datasets.
However, DEIQT still requires end-to-end fine-tuning of a transformer model, leading to increased training times.

\begin{figure*}[hbt!]
\centering
\includegraphics[trim={0.5cm 0.7cm 1.4cm 0.5cm}, clip, scale=0.6]{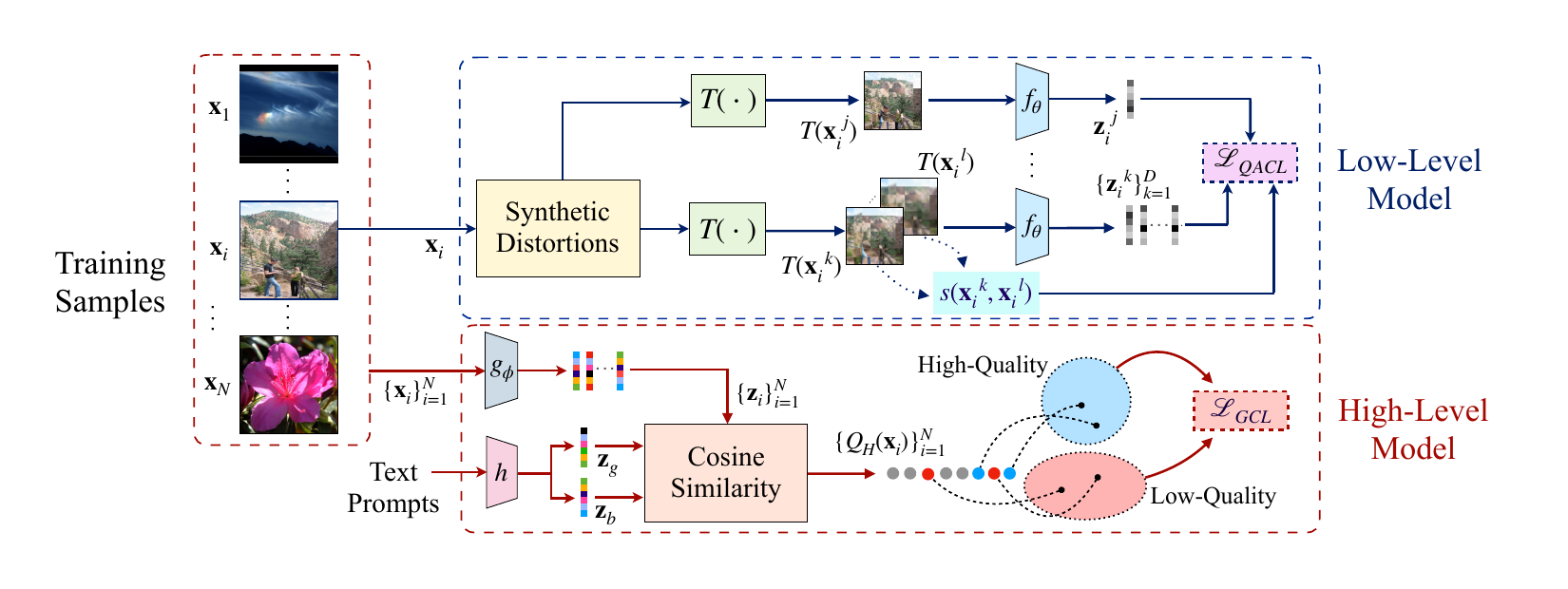}
\caption{Illustration of the GRepQ framework. The low-level model, $f_{\theta}$, is trained using multiple distorted versions of an image $\mathbf{x}_i$ subjected to the fragment sampling operation $T(\cdot)$. $\mathbf{x}_i^j$ denotes an anchor image. The perceptual similarity measure $s(\cdot, \cdot)$ is used to weigh the feature similarities of pairs of distorted images $\mathbf{x}_i^k$ and $\mathbf{x}_i^l$ in \cref{qacl}. The high-level model, $g_{\phi}$, is trained using \cref{loss_gcl} after selecting groups of features based on their cosine similarities with embeddings $\mathbf{z}_g$ and $\mathbf{z}_b$ of antonym text prompts that relate to higher and lower quality respectively. Embeddings are obtained from the text-encoder $h(\cdot).$}
\label{fig:GRepQ}
\end{figure*}

\section{Method}\label{sec:method}

We first describe our approach to learning generalizable low-level and high-level quality representations. The overall framework is illustrated in \cref{fig:GRepQ}. We discuss how quality is predicted in the data-efficient and zero-shot settings. 

\subsection{Low-Level Representation Model}\label{sec:method_llm}

Contrastive learning for image quality \cite{contrique, reiqa, qap} discriminates images based on varied types and levels of distortions to capture low-level information. While this provides very good pre-trained feature encoders without learning from human labels, images with different distortion types are often treated as a negative pair with respect to an anchor image, and the features of such images are pulled apart. This leads to two main issues. An image with a different distortion type may have a similar perceptual quality as the anchor. Secondly, since image representations with different distortion types are separated, this hurts the model's generalizability to represent unseen distortions. The goal of our work is to address both these limitations through a quality-aware contrastive learning loss.

We introduce a novel quality-aware contrastive loss, where positive and negative pairs (pairs of images considered similar and dissimilar in quality, respectively) are selected based on their perceptual similarity. This allows a soft weighting such that a similarity weight close to one treats the pair of images as positive and pulls their corresponding representations closer.
Similarly, image features are pulled apart when the perceptual similarity is near zero. This differs from the way prior methods use contrastive learning. In particular, our framework allows the selection of pairs regardless of their distortions, allowing for generalization. 

\noindent \textbf{Image Augmentation:} In order to train the feature encoder using contrastive learning, we generate multiple synthetically distorted versions of a camera-captured image and sample fragments from each image. Four synthetic distortions are generated: blur, compression, noise, and color saturation, at two levels each. Fragment sampling has proven effective in retaining the global quality information of an image \cite{fastvqa}. To obtain fragments, we divide an image into grids, and random mini-patches are extracted from each of the grid locations. The mini-patches are then stitched together to yield a single fragmented image that is used to train the model. An augmentation is generated by randomly sampling another set of mini-patches from the same image to obtain another fragmented image. Note that this augmentation is quality preserving and can be used as a hard-positive pair in contrastive loss. 

\noindent \textbf{Quality-Aware Contrastive Loss:} 
We contrast multiple distorted versions of the same scene to learn quality representations and mitigate content bias. 
Consider a batch of $N_b$ images $\{\mathbf{x}_i\}_{i=1}^{N_b}$, where each image has $D$ distorted versions. Let $\mathbf{x}_i^j$ and $\mathbf{x}_i^k$ denote two distorted versions of an image $\mathbf{x}_i$ where $j,k\in\{1,2,\ldots,D\}$. Let $\mathbf{z}_i^j$ and $\mathbf{z}_i^k$ be the respective unit-norm feature representations obtained as $\mathbf{z}_i^j = f_{\theta}(T(\mathbf{x}_i^j))$ and $\mathbf{z}_i^k = f_{\theta}(T(\mathbf{x}_i^k))$, where $T(\cdot)$ is the fragment sampling operation and $f_\theta(\cdot)$ is the feature encoder. Let $s(\cdot, \cdot) : \mathbb{R}^{M \times N} \times \mathbb{R}^{M \times N} \rightarrow [0, 1]$ denote a perceptual similarity measure between two images with the same content. Further, let $p_{\tau}(\mathbf{z}_1, \mathbf{z}_2) = \text{exp}(\mathbf{z}_1 \cdot \mathbf{z}_2 / \tau)$.  We overcome the limitation of existing contrastive learning methods which require hard positives and negatives through the above soft similarity measure to label positives and negatives. The similarity measures the closeness of distorted versions in terms of intrinsic quality attributes and provides a confidence weight in contrastive loss. 

Our quality-aware contrastive loss is given by $\mathcal{L}_{QACL} = \sum_{i=1}^{N_b} \sum_{j=1}^D \mathcal{L}_{i}^j$, where $\mathcal{L}_i^j$ is given by
\begin{equation}
    \mathcal{L}_i^j = - \text{log} \frac{p_{\tau_1}(\mathbf{z}_i^j, \mathbf{z}_{i}^{j+}) + \sum_{k \neq j} s(\mathbf{x}_i^j, \mathbf{x}_i^k) p_{\tau_1}(\mathbf{z}_i^j, \mathbf{z}_i^k)}{p_{\tau_1}(\mathbf{z}_i^j, \mathbf{z}_{i}^{j+}) + \sum_{k \neq j} p_{\tau_1}(\mathbf{z}_i^j, \mathbf{z}_i^k)},
    \label{qacl}
\end{equation}
where $\mathbf{z}_i^{j+}$ is the representation of an augmentation of the image $\mathbf{x}_i^j$, and $\tau_1$ is a temperature hyperparameter. Note that since $p_{\tau_1}(\mathbf{z}_i^j, \mathbf{z}_i^k)=s(\mathbf{x}_i^j, \mathbf{x}_i^k)p_{\tau_1}(\mathbf{z}_i^j, \mathbf{z}_i^k)+(1-s(\mathbf{x}_i^j, \mathbf{x}_i^k))p_{\tau_1}(\mathbf{z}_i^j, \mathbf{z}_i^k)$, $\mathbf{x}_i^k$ is treated as similar to $\mathbf{x}_j^k$ with weight $s(\mathbf{x}_i^j, \mathbf{x}_i^k)$ and dissimilar with weight $(1-s(\mathbf{x}_i^j, \mathbf{x}_i^k))$. 
The similarity function makes the learning distortion type agnostic since it measures relative degradation without knowledge of the distortion type, making the learned features generalizable to different (and unseen) distortions. The InfoNCE \cite{infonce} loss can be seen as a special case of $\mathcal{L}_{QACL}$ when $s(\cdot, \cdot) = 0$.

It is desirable that the perceptual similarity measure used satisfies a few properties: ($1$) It captures intrinsic quality-specific attributes, such as structure, sharpness, or contrast, ($2$) It is capable of handling various distortion types used during training and correlates well with human judgments on these distortions, ($3$) It captures local and global quality information relevant to the human visual system, and $(4)$ It predicts similarity with fairly low complexity to enable faster training times. We explore different similarity measures such as FSIM \cite{fsim}, SSIM \cite{ssim}, GMSD \cite{gmsd}, MS-SSIM\cite{msssim} and LPIPS\cite{lpips} in our work. Degraded reference IQA \cite{degraded} has also shown that such similarity measures can be used to distinguish between distorted versions of a degraded reference image. Finally, we note that such similarity measures can be used to compare different distorted versions of an anchor image since all of them have the same content. Thus, we do not include variations of different images in this loss.

\subsection{High-Level Representation Model}\label{sec:method_hlm}

To understand the scene context for IQA, we adapt the CLIP model to the IQA task. Image quality can be obtained using CLIP by measuring the cosine similarity between the image feature and the text embeddings of a pair of antonym prompts such as \texttt{[``a good photo.'', ``a bad photo.'']} \cite{clipiqa}. Although CLIP has a reasonable zero-shot quality prediction performance in terms of correlation with human opinion, the representations are not specifically crafted for the task of IQA, leading to a performance gap. We bridge this gap by fine-tuning the image encoder of CLIP through an unsupervised loss, as described below. 

\noindent \textbf{Contrastive Learning Over Groups:} Analyses of vision-language models show that text representations are richer than image representations \cite{textisgood1, textisgood2}. Thus, we fix the text encoder in the CLIP model and only update the image encoder for the IQA task. The text representations corresponding to the antonym prompts remain the same during training and testing. We propose a loss for updating the image encoder that aims at separating images in a batch into groups based on how close the image representations are to two text-prompt embeddings. We then seek to align the representations of images within each group and separate the representations across groups. Such a loss simultaneously ensures that the intra-group feature entropy (entropy of representations within each group) is minimized and the inter-group entropy (entropy of features between groups) is maximized \cite{entropy, scl}.    

Consider a batch consisting of $N$ images $\{\mathbf{x}_{i}\}_{i=1}^N$ with visual representations $\{\mathbf{z}_{i}\}_{i=1}^N$. The representations are obtained as $\mathbf{z}_i = g_{\phi}(\mathbf{x}_i)$, where $g_{\phi}(\cdot)$ is the CLIP image encoder. Let $\mathbf{z}_g$ and $\mathbf{z}_b$ correspond to the prompt representations of \texttt{``a good photo.''} and \texttt{`` bad photo.''} respectively. We construct two groups of images, $\mathcal{S}_g$ and $\mathcal{S}_b$, that correspond to higher and lower quality respectively based on the quality estimated as 
\begin{equation}
    Q_H (\mathbf{x}_i) = \frac{1}{ 1 + \text{exp}(k_2(\mathbf{z}_i \cdot \mathbf{z}_b - \mathbf{z}_i \cdot \mathbf{z}_g))}, 
    \label{qual_c}
\end{equation}
where $k_2$ is a scaling parameter. Let the features $\{\mathbf{z}_{i}\}_{i=1}^N$ sorted in increasing order of $Q_H$ be $\{ \mathbf{z}_{(1)}, \mathbf{z}_{(2)}, \cdots , \mathbf{z}_{(N)} \}$. We obtain the groups as $\mathcal{S}_b = \{z_{(i)} \}_{i=1}^{M} $ and $\mathcal{S}_g = \{z_{(i)} \}_{i=N-M+1}^{N} $, where $M = \texttt{round}(N/k)$, $k$ is a hyperparameter that decides the separability of lower and higher quality groups within a batch of images, and $M$ denotes the group size. Let $\mathcal{S}_b(i) = \mathcal{S}_b \setminus \{\mathbf{z}_i\}$ and $\mathcal{S}_g(i) = \mathcal{S}_g \setminus \{\mathbf{z}_i\}$. Our group contrastive loss used for fine-tuning is expressed as 
\begin{align}
    \nonumber \mathcal{L}_{GCL} = & - \sum_{\mathbf{z}_i \in \mathcal{S}_g} \text{log} \frac{ \sum_{\mathbf{z}_j \in \mathcal{S}_g(i) }p_{\tau_2}(\mathbf{z}_i, \mathbf{z}_j)}{ \sum_{\mathbf{z}_j \in \mathcal{S}_g(i) \cup \mathcal{S}_b } p_{\tau_2}(\mathbf{z}_i, \mathbf{z}_j)} \\
    & - \sum_{\mathbf{z}_i \in \mathcal{S}_b} \text{log} \frac{ \sum_{\mathbf{z}_j \in \mathcal{S}_b(i) }p_{\tau_2}(\mathbf{z}_i, \mathbf{z}_j)}{ \sum_{\mathbf{z}_j \in \mathcal{S}_b(i) \cup \mathcal{S}_g } p_{\tau_2}(\mathbf{z}_i, \mathbf{z}_j)}. 
    \label{loss_gcl}
\end{align}

While creating groups, a quality separation gap and closeness of quality scores of images in each group are necessary for effective contrastive learning. 
The parameter $k$ controls this separability and is a hyperparameter that needs to be appropriately chosen. 

\subsection{Mapping Representations to Objective Quality}\label{sec:rel_GRepQ}

\noindent \textbf{Data-Efficient Quality Prediction:} Once the high and low-level features are learned, they are concatenated and regressed with mean opinion scores on the evaluation datasets using a few samples from each dataset. We use a linear SVR $f_d(\cdot): \mathbb{R}^P \rightarrow \mathbb{R}$ on features of target datasets, where $P$ is the feature dimension. The data-efficient quality of any new image $\mathbf{x}$ can simply be computed using its corresponding feature representation $\mathbf{z}_\mathbf{x} \in \mathbb{R}^P$ as  
\begin{equation}
    \mathbf{GRepQ}_{D}(\mathbf{x}) = f_d(\mathbf{z}_\mathbf{x}).   
    \label{GRepQ_d}
\end{equation}
Our approach offers the advantage of requiring no end-to-end training using the limited labels on a new target database. 

\begin{figure}
    \centering
    \includegraphics[trim={22cm 7cm 22cm 7cm}, clip, scale=0.35]{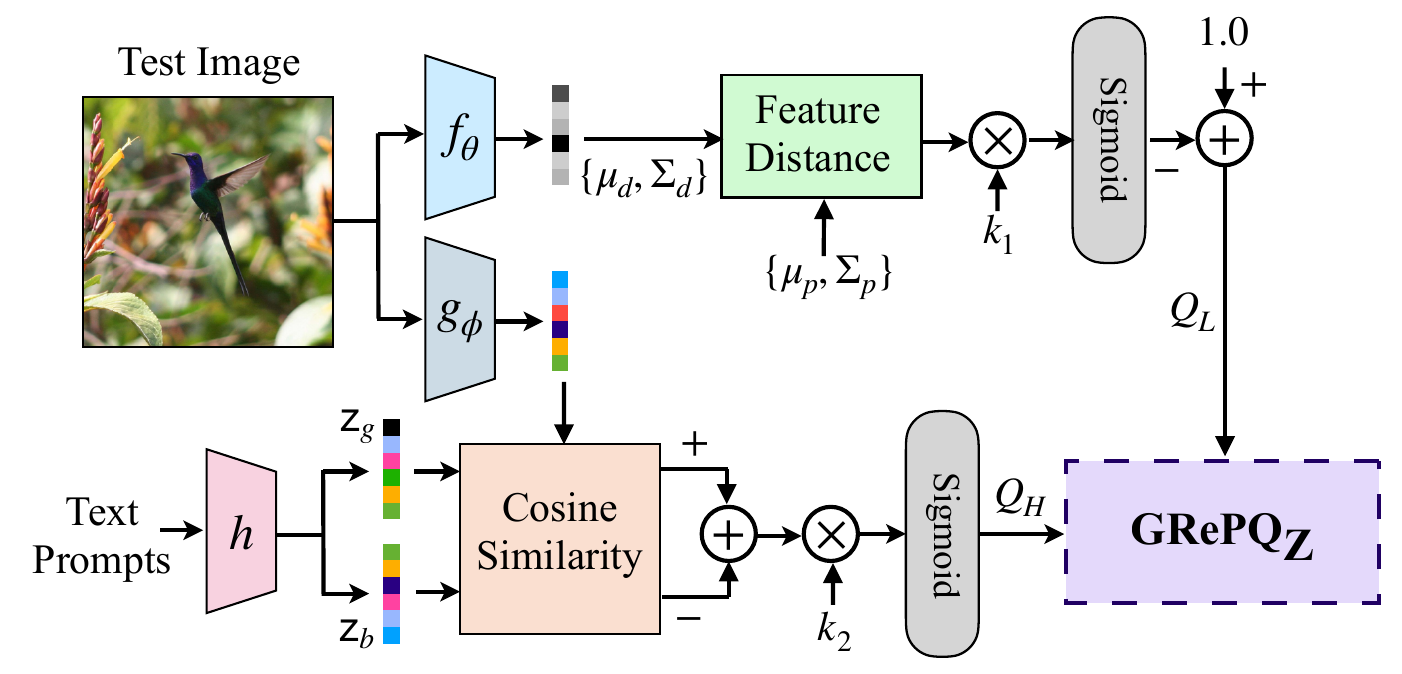}
    \caption{$\text{GRepQ}_\text{Z}$: Computing zero-shot image quality. The text-prompts used for evaluation are \texttt{"a good/bad photo."}.}
    \label{fig:GRepQ_eval}
\end{figure}

\noindent \textbf{Zero-Shot Quality Prediction:} We use different approaches for the low-level and high-level representations to predict quality without using any supervision. For the low-level features,  we compute a distance between the features of the input image and that of a corpus of pristine images similar to NIQE as 
\begin{equation}
    d(\mathbf{x}) = \sqrt{(\mu_p - \mu_d)^T \bigg( \frac{\Sigma_p + \Sigma_d}{2} \bigg)^{-1} (\mu_p - \mu_d) },
    \label{niqe_distance}
\end{equation}
where $\mu_p$ and $\Sigma_p$ are the mean and covariance of the representations from the low-level encoder corresponding to patches of pristine images. $\mu_d$ and $\Sigma_d$ are the mean and covariance of the representations of the patches from an image $\mathbf{x}$. Here, non-overlapping patches of size $R \times R$ are extracted from the image to estimate the relevant statistics of the features. The low-level quality is then predicted as 
\begin{equation}
    Q_L(\mathbf{x}) = \frac{1}{1 + \text{exp}( k_1 d(\mathbf{x}))}, 
    \label{qd_align}
\end{equation}
where $k_1$ is a scaling parameter. The quality from the high-level representations can be predicted using \cref{qual_c}. The overall image quality is then measured as 
\begin{equation}
    \text{$\mathbf{GRepQ}_Z$}(\mathbf{x}) = Q_H(\mathbf{x}) + Q_L(\mathbf{x}),
    \label{GRepQ_eq}
\end{equation}
and is illustrated in \cref{fig:GRepQ_eval}.

\section{Experiments}\label{sec:experiments}

\subsection{Training and Implementation Details}\label{sec:experiments_traineval}

\noindent \textbf{Training Dataset for Representation Learning:} We train the low and high-level feature encoders on the FLIVE dataset \cite{paq2piq} using a subset of $5000$ real-world images encompassing a variety of authentic distortions with different resolutions and aspect ratios. The diverse content and distortions make it conducive to learning representations that can be generalized to diverse images. No human annotations were used during this training. 

\noindent \textbf{Low-Level Encoder:} We use a ResNet18 (performance with a Resnet50 was found to be similar) without pre-trained weights for the low-level feature encoder. The contrastive loss in  \cref{qacl} is trained by projecting the features from the penultimate layer of ResNet18 onto $\mathbb{R}^{128}$. Images are fragmented into $7 \times 7$ grids, and random mini-patches from each grid location are stitched together to form $224 \times 224$ sized patches. The temperature $\tau_1$ is fixed at $0.5$. A batch consists of $8$ images with $8$ distorted versions each. The model is trained for $15$ epochs using the AdamW \cite{adamw} optimizer with a weight decay of $0.05$ and an initial learning rate of $10^{-4}$. A cosine learning rate scheduler is used. To guide the quality-aware contrastive training, we employ FSIM as the perceptual similarity measure. 

\begin{table*}[!ht]
\adjustbox{max width=\textwidth}
\centering
\resizebox{\textwidth}{!}{%
\begin{tabular}{c|c|ccc|ccc|ccc|ccc|ccc}
\hline
Method & \multirow{2}{*}{\begin{tabular}[c]{@{}c@{}}Method\\ Type\end{tabular}} & \multicolumn{3}{c|}{CLIVE} & \multicolumn{3}{c|}{KonIQ} & \multicolumn{3}{c|}{CSIQ} & \multicolumn{3}{c|}{LIVE} & \multicolumn{3}{c}{PIPAL} \\ \cline{1-1} \cline{3-17} 
Labels &  & 50 & 100 & 200 & 50 & 100 & 200 & 50 & 100 & 200 & 50 & 100 & 200 & 50 & 100 & 200 \\ \hline
TReS\cite{transformers1} & \multirow{5}{*}{\begin{tabular}[c]{@{}c@{}}End-to-\\ end\\ Fine\\ Tuning\end{tabular}} & 0.670 & 0.751 & 0.799 & 0.713 & 0.719 & 0.791 & 0.791 & 0.811 & 0.878 & 0.901 & 0.927 & \textbf{0.957} & 0.186 & 0.349 & 0.501 \\
HyperIQA\cite{hyperiqa} &  & 0.648 & 0.725 & 0.790 & 0.615 & 0.710 & 0.776 & 0.790 & 0.824 & 0.909 & 0.892 & 0.912 & 0.929 & 0.102 & 0.302 & 0.379 \\
DEIQT\cite{deiqt} &  & 0.667 & 0.718 & 0.812 & 0.638 & 0.682 & 0.754 & 0.821 & 0.891 & \textbf{0.941} & 0.920 & \textbf{0.942} & 0.955 & 0.396 & 0.410 & 0.436 \\
MANIQA\cite{maniqa} &  & 0.642 & 0.769 & 0.797 & 0.652 & 0.755 & 0.810 & 0.794 & 0.847 & 0.874 & 0.909 & 0.928 & \textbf{0.957} & 0.136 & 0.361 & 0.470 \\
LIQE\cite{liqe} &  & 0.691 & 0.769 & 0.810 & 0.759 & 0.801 & 0.832 & 0.838 & 0.891 & 0.924 & 0.904 & 0.934 & 0.948 & - & - & - \\ \hline
Resnet50\cite{resnet} & \multirow{7}{*}{\begin{tabular}[c]{@{}c@{}}Simple\\ Feature\\ Regre-\\ ssion\end{tabular}} & 0.576 & 0.611 & 0.636 & 0.635 & 0.670 & 0.707 & 0.793 & 0.890 & 0.935 & 0.871 & 0.906 & 0.922 & 0.150 & 0.220 & 0.302 \\
CLIP\cite{clip} &  & 0.664 & 0.721 & 0.733 & 0.736 & 0.770 & 0.782 & 0.841 & 0.892 & 0.941 & 0.896 & 0.923 & 0.941 & 0.254 & 0.303 & 0.368\\
CONTRIQUE\cite{contrique} &  & 0.695 & 0.729 & 0.761 & 0.733 & 0.794 & 0.821 & 0.840 & \textbf{0.926} & 0.940 & 0.891 & 0.922 & 0.943 & 0.379 & 0.437 & 0.488 \\
Re-IQA\cite{reiqa} &  & 0.591 & 0.621 & 0.701 & 0.685 & 0.723 & 0.754 & \textbf{0.893} & 0.907 & 0.923 & 0.884 & 0.894 & 0.929 & 0.280 & 0.350 & 0.431 \\ \cline{1-1} \cline{3-17} 
$\text{GRepQ}_\text{D}$ (LL) &  & 0.531 & 0.565 & 0.613 & 0.620 & 0.647 & 0.679 & 0.794 & 0.805 & 0.832 & 0.866 & 0.880 & 0.886 & 0.395 & 0.410 & 0.431 \\
$\text{GRepQ}_\text{D}$ (HL) &  & 0.740 & 0.770 & 0.796 & 0.794 & 0.813 & 0.843 & 0.869 & 0.905 & 0.932 & 0.904 & 0.927 & 0.944 & 0.410 & 0.415 & 0.427 \\
$\text{GRepQ}_\text{D}$ (HL + LL) &  & \textbf{0.760} & \textbf{0.791} & \textbf{0.822} & \textbf{0.812} & \textbf{0.836} & \textbf{0.855} & 0.878 & \multicolumn{1}{l}{0.914} & \textbf{0.941} & \textbf{0.926} & 0.937 & 0.953 & \textbf{0.489} & \textbf{0.518} & \textbf{0.548} \\ \hline
\end{tabular}}
\caption{SRCC performance comparison of $\text{GRepQ}_\text{D}$ with other NR-IQA methods trained using few labels on various IQA databases. The methods are segregated into end-to-end trained (top five) and feature-learning-based (next four) methods. LL and HL correspond to low and high-level models respectively. The best-performing methods are bolded.}
\label{table:main_table_de}
\end{table*}

\noindent \textbf{High-Level Encoder:} We fine-tune CLIP's image encoder while keeping the text encoder fixed. The image encoder consists of a Resnet50 backbone with an additional attention-pooling layer. To enable contrastive learning over groups specified in \cref{loss_gcl}, a projection head is used to contrast features in $\mathbb{R}^{128}$. The images are center-cropped to a size of $224 \times 224$, and a batch size of $N = 128$ is used. Based on the coarse predictions obtained using \cref{qual_c}, we use a separability hyperparameter $k=8$ to divide the batch of images into groups of size $M = 16$. Once the groups are formed, the image encoder is trained using \cref{loss_gcl} with a temperature $\tau_2 = 0.1$. The model is trained for $15$ epochs using an Adam optimizer with an initial learning rate of $5 \times 10^{-6}$. The scaling parameter $k_2$ is set to $10$. 

\noindent \textbf{Zero-Shot Quality Prediction using Low-Level Encoder:} For the zero-shot quality prediction using \cref{niqe_distance}, we select $125$ pristine image patches as used in literature \cite{clmi} (chosen based on sharpness and colorfulness). Patches of size $96 \times 96$ are extracted from the pristine images and the test image. The scaling parameter $k_1$ is set to $0.01$. 

All the implementations were done in PyTorch using two 11GB Nvidia GeForce RTX 2080 Ti GPUs.  

\subsection{Experimental Setup}

We present the details of the two main evaluation settings: data-efficient setting and the zero-shot setting. 
In the data-efficient setting, we train our data-efficient framework $\text{GRepQ}_\text{D}$, using a few samples from each evaluation dataset.
We randomly split each evaluation dataset into 80\% and 20\% and use the 20\% subset for testing. We select a random subset of 50, 100, or 200 samples from the 80\% for training a linear support vector regressor (SVR) on the features. 
We use Spearman's rank order correlation coefficient (SRCC) between the objective and subjective scores to evaluate the models' performance. 
We report the median performance obtained across 10 splits of each evaluation dataset. 
The results with respect to Pearson's linear correlation coefficient (PLCC) are given in the supplementary. 
In the zero-shot setting, no training on any evaluation dataset is required, and we test on the entire evaluation dataset. 
 
\noindent \textbf{Evaluation Datasets:} We choose a variety of datasets spanning different types of distortions to demonstrate the effectiveness of our framework for the three experimental settings. 
Since the training images are sampled from the FLIVE dataset, we do not evaluate them on FLIVE. 
We evaluate two popular in-the-wild datasets: CLIVE \cite{clive}, KONiQ \cite{koniq}, and three synthetic or processed image datasets: LIVE-IQA\cite{liveiqa},  CSIQ\cite{csiq} and PIPAL\cite{pipal}. \textbf{CLIVE} contains $1,162$ images captured from multiple mobile devices. \textbf{KonIQ-10K} contains $10073$ in-the-wild images. \textbf{LIVE-IQA}\cite{liveiqa} contains $29$ scenes along with $779$ distorted images containing JPEG compression, blur, noise, and fast-fading distortions. CSIQ\cite{csiq} consists of $30$ original images with $866$ distorted images with blur, contrast, and JPEG compression distortions. \textbf{PIPAL} is a large IQA database consisting of $23,200$ images with $40$ different distortions per image, including GAN-generated artifacts, making this dataset very challenging to evaluate. 

\subsection{Data-Efficient Setting}\label{sec:data_efficient_exp}

We compare $\text{GRepQ}_\text{D}$ with other state-of-the-art (SoTA) end-to-end NR-IQA methods: TReS\cite{transformers1}, HyperIQA\cite{hyperiqa}, and MANIQA\cite{maniqa}, the data-efficient method DEIQT\cite{deiqt}, and feature based methods: Resnet50\cite{resnet}, CLIP\cite{clip}, CONTRIQUE\cite{contrique} and Re-IQA\cite{reiqa}. We note that LIQE\cite{liqe} is not trainable on PIPAL and thus its entry is left blank. For the methods requiring feature regression, the SVR parameters are optimized to yield the best performances. 
To ensure fair comparisons, the median performance of all methods over ten train-test splits are reported.

\cref{table:main_table_de} presents comparisons on the data-efficient training of $\text{GRepQ}_\text{D}$ against other NR-IQA methods. The results indicate that $\text{GRepQ}_\text{D}$ outperforms other methods on all datasets in almost all three data regimes (50, 100, and 200 samples).  We notice that $\text{GRepQ}_\text{D}$ outperforms even end-to-end trained models despite using a simple SVR. The superior performance over Re-IQA, which may also be considered as an ensemble of two sets of features, demonstrates the superiority of both our low and high level features. While it may appear that the high-level model performs better than the low-level model in most of the scenarios, we provide examples in  \cref{sec:experiments_ablations}, where the low-level model could also be more accurate. Thus, there is a need for both the high and low-level representations. 
As an extreme case, we also present results in the fully-supervised setting in the supplement. 

\begin{table}[!t]
\adjustbox{max width=\columnwidth}
\centering
\resizebox{\columnwidth}{!}{%
\begin{tabular}{c|c|c|c|c|c}
\hline
Method & CLIVE & KonIQ & CSIQ & LIVE & PIPAL\\ 
\hline
NIQE\cite{niqe} & 0.463 & 0.530 & 0.613 & 0.836 & 0.153 \\
IL-NIQE\cite{ilniqe} & 0.440 & 0.507  & \textbf{0.814} & \textbf{0.847} & 0.282\\
CL-MI\cite{clmi} & 0.507 & 0.645  & 0.588 & 0.663 & 0.303 \\
CLIP-IQA\cite{clipiqa} & 0.612 & 0.700  & 0.690 & 0.652 & 0.261\\ \hline
$\text{GRepQ}_\text{Z}$ & \textbf{0.740} & \textbf{0.768} & 0.693 & 0.741 &  \textbf{0.436}\\ \hline
\end{tabular}}
\caption{Performance comparison of $\text{GRepQ}_\text{Z}$ (zero-shot) with other zero-shot methods on various IQA databases.}
\label{table:main_table_zs}
\end{table}

\begin{table}[!t]
\adjustbox{max width=\columnwidth}
\centering
\resizebox{\columnwidth}{!}{%
\begin{tabular}{c|cc|c|c|c|c}
\hline
Training & \multicolumn{2}{c|}{FLIVE}                           & KonIQ          & CLIVE          & LIVE           & CSIQ           \\ \hline
Testing  & \multicolumn{1}{c|}{CLIVE}          & KonIQ          & CLIVE          & KonIQ          & CSIQ           & LIVE           \\ \hline
HyperIQA & \multicolumn{1}{c|}{0.758}          & 0.735          & 0.785          & 0.772          & 0.744          & 0.926          \\
TReS     & \multicolumn{1}{c|}{0.713}          & 0.740          & 0.786          & 0.733          & 0.761          & \textbf{-}     \\
CONTRIQUE   & \multicolumn{1}{c|}{0.710}          &  0.781         &  0.731         &  0.676    &  \textbf{0.823}    & 0.925                     \\
DEIQT    & \multicolumn{1}{c|}{0.733}          & 0.781          & \textbf{0.794} & 0.744          & 0.781 & \textbf{0.932} \\ \hline
$\text{GRepQ}_\text{C}$  & \multicolumn{1}{c|}{\textbf{0.774}} & \textbf{0.815} & 0.774          & \textbf{0.792} & 0.770          & 0.893          \\ \hline
\end{tabular}}
\caption{Cross-dataset performance of $\text{GRepQ}_\text{C}$ along with other NR-IQA methods. Results for methods apart from CONTRIQUE are from \cite{deiqt}.}
\label{table:cross_dataset}
\end{table}

\subsection{Zero-Shot Setting}\label{sec:experiments_zs}
 
Since zero-shot methods are trained without human supervision, we compare $\text{GRepQ}_\text{Z}$ with unsupervised or completely blind NR-IQA methods such as NIQE \cite{niqe}, IL-NIQE \cite{ilniqe}, contrastive learning with mutual information (CL-MI) \cite{clmi}, and CLIP-IQA \cite{clipiqa}. We utilize entire evaluation databases for testing all the methods. 

\cref{table:main_table_zs} shows that $\text{GRepQ}_\text{Z}$ consistently outperforms other methods on three out of five datasets by considerable margins. 
$\text{GRepQ}_\text{Z}$ achieves SoTA performance even on the challenging PIPAL dataset, containing diverse distortions, particularly images restored by various restoration (including GAN-based) methods for super-resolution and denoising. 
A $44\%$ improvement in SRCC is shown over the second-best-performing algorithm (CL-MI). The lower performance of $\text{GRepQ}_\text{Z}$ on LIVE and CSIQ is attributed to content bias of both the low-level and high-level models. Although the low-level model is trained in a content conditional manner, the features perhaps do suffer from some residual content bias. Since LIVE and CSIQ contain very few unique scene content, the residual content bias leads to reduced performance of our zero-shot model. 
Despite these challenges, $\text{GRepQ}_\text{Z}$ still achieves competitive performance, showing its generalization capability in the zero-shot setting. 

\begin{table}[!t]
\centering
\begin{tabular}{c|c|c|c}
\hline
Similarity Measure  & $50$ & $100$ & $200$ \\ \hline
None  & 0.381 & 0.413 & 0.452  \\
SSIM  & 0.533 & 0.558 & 0.590  \\
MS-SSIM & 0.527 & 0.561 &  0.575 \\
GMSD  & 0.544 & 0.570 & 0.583  \\ 
LPIPS & 0.578 & 0.605 & 0.629  \\\hline
\textbf{FSIM} & \textbf{0.620} & \textbf{0.647} & \textbf{0.679} \\ \hline
\end{tabular}
\caption{SRCC performance analysis on the KonIQ dataset of the impact of different perceptual similarity measures on the low-level model under the data-efficient setting.}
\label{table:measures}
\end{table}

\subsection {Cross-Database Experiments} 
We also show the effectiveness of our features through cross-database experiments. Here, a single linear SVR (ridge regressor) is trained on an entire dataset and tested on other intra-domain datasets in the authentic and synthetic image settings.
The results in \cref{table:cross_dataset} indicate that
$\text{GRepQ}_\text{C}$ (defined as the cross-dataset prediction evaluated using \cref{GRepQ_d}) achieves competitive (also best) performances in most of the evaluation settings. 

\subsection{A Deeper Understanding of GRepQ Features}\label{sec:experiments_ablations}

\noindent \textbf{Choice of Perceptual Similarity Measures in the Low-Level Feature Encoder:} We compare different popular perceptual similarity measures such as SSIM \cite{ssim}, MS-SSIM\cite{msssim}, FSIM\cite{fsim}, LPIPS\cite{lpips} and GMSD\cite{gmsd} used in the low-level feature encoder in \cref{table:measures}. The low-level encoders are trained using these measures under similar training settings. We also train an encoder without any similarity measure (denoted by None) to show a need for quality-aware contrastive learning. In this case, all the other distorted versions of an image are treated as negatives, while the augmented version is the only positive. 
In \cref{table:measures}, we show the low-level encoder's data-efficient performances on KonIQ. We see that FSIM outperforms all other measures. We note that the superior performance of FSIM in this context is consistent with its superior performance as an FR-IQA metric across multiple datasets. \\

\begin{figure}[!t]
\centering
\begin{subfigure}[b]{0.48\columnwidth}
   \includegraphics[width=\columnwidth]{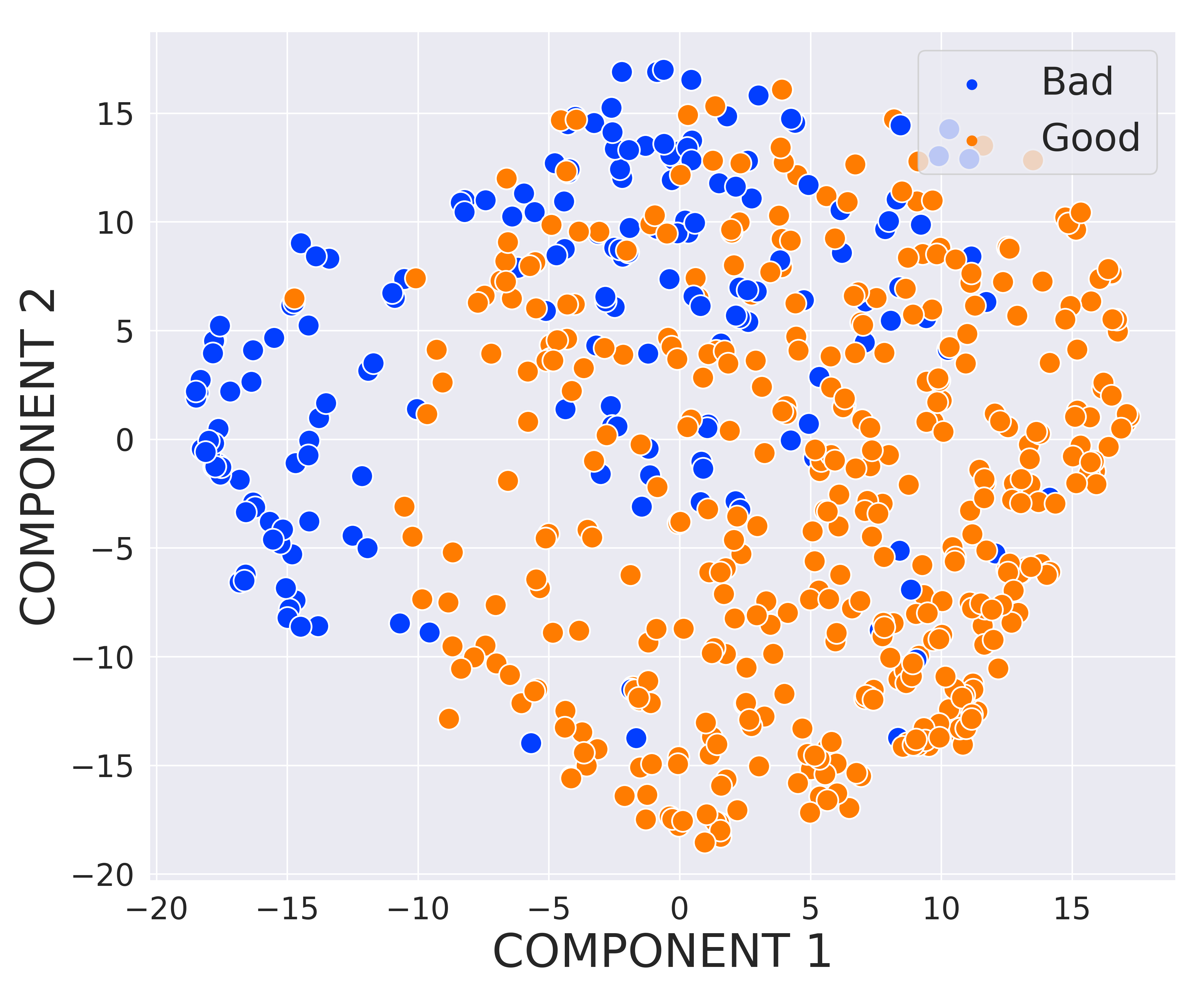}
   \caption{}
   \label{fig:tsne_clip} 
\end{subfigure}
\hfill
\begin{subfigure}[b]{0.48\columnwidth}
   \includegraphics[width=\columnwidth]{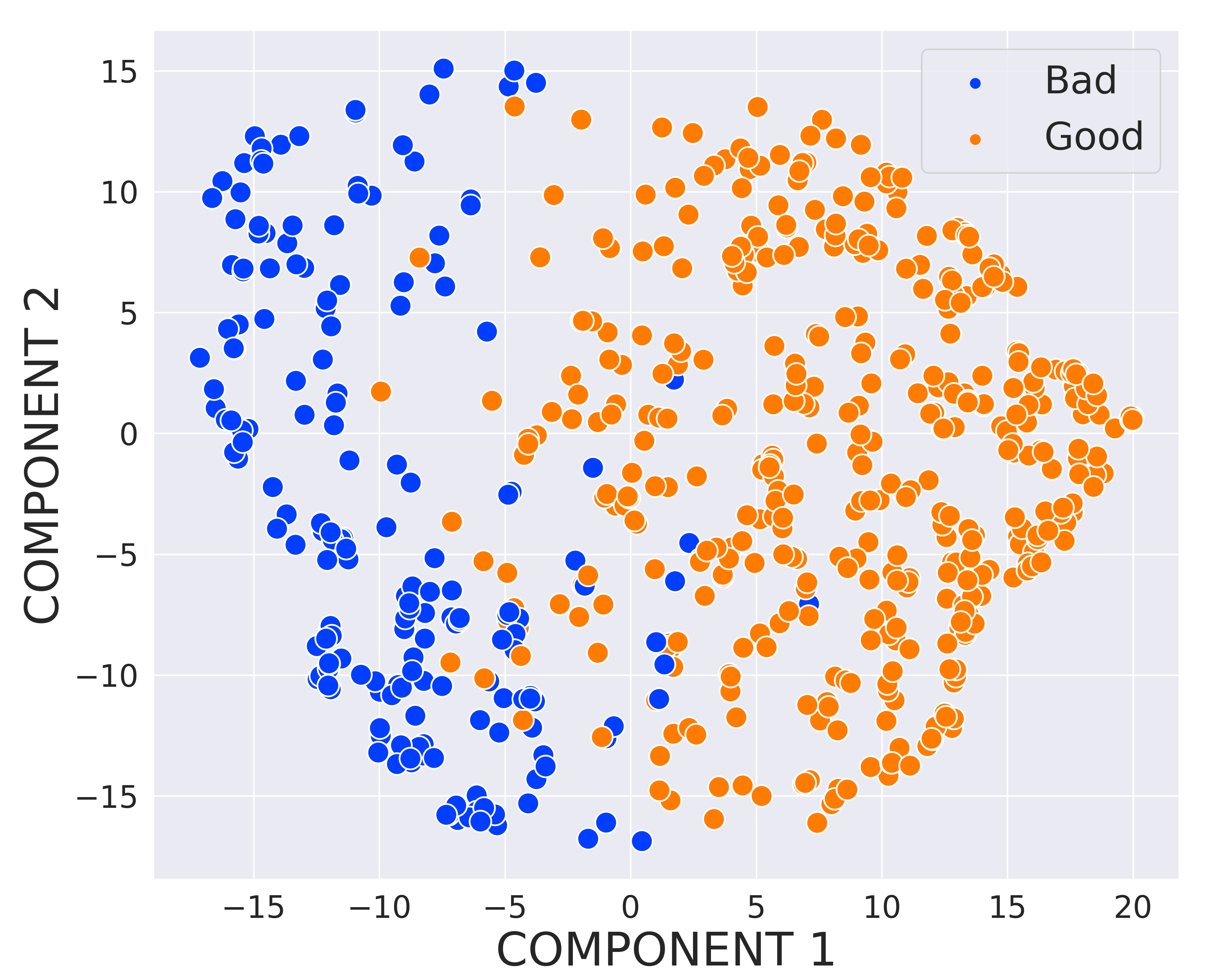}
   \caption{}
   \label{fig:tsne_gcl}
\end{subfigure}
\caption{A visualization of feature representations of images from a combination of CLIVE and the test set of KonIQ using t-SNE for features from (a) the zero-shot CLIP image encoder and (b) our fine-tuned high-level encoder. Blue and orange points correspond to bad and good-quality images respectively.}
\label{fig:tsne}
\end{figure}

\begin{table*}[!ht]
\begin{tabular}{ccccc} \\
\includegraphics[width=0.19\textwidth]{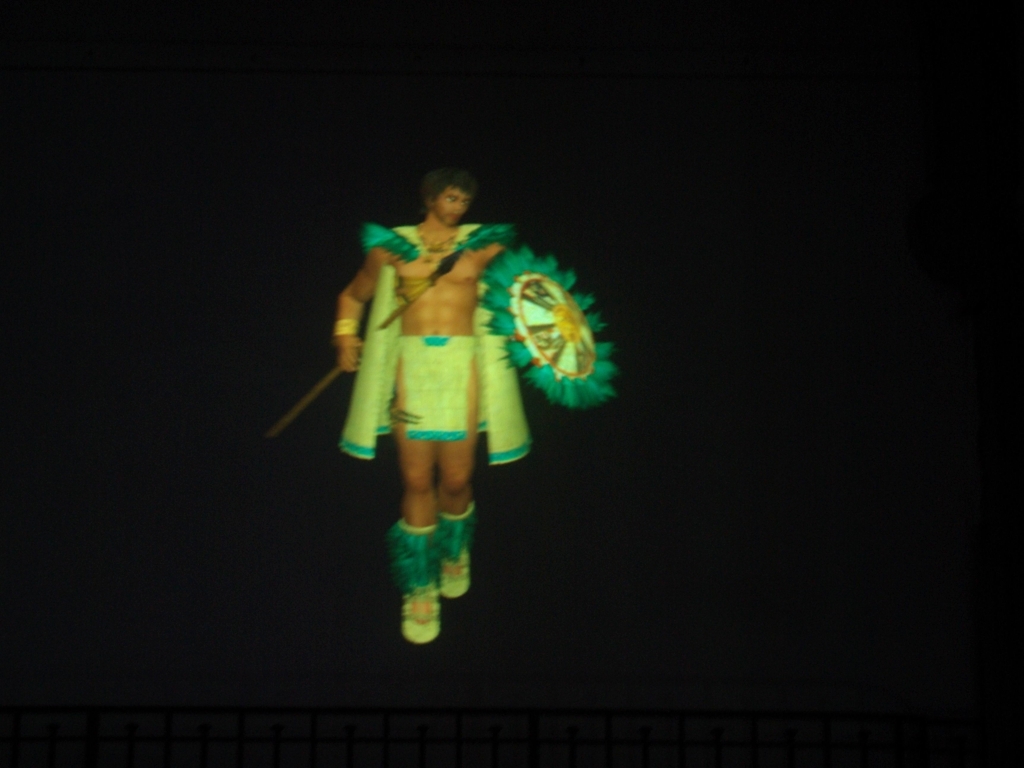}     &    \includegraphics[width=0.19\textwidth]{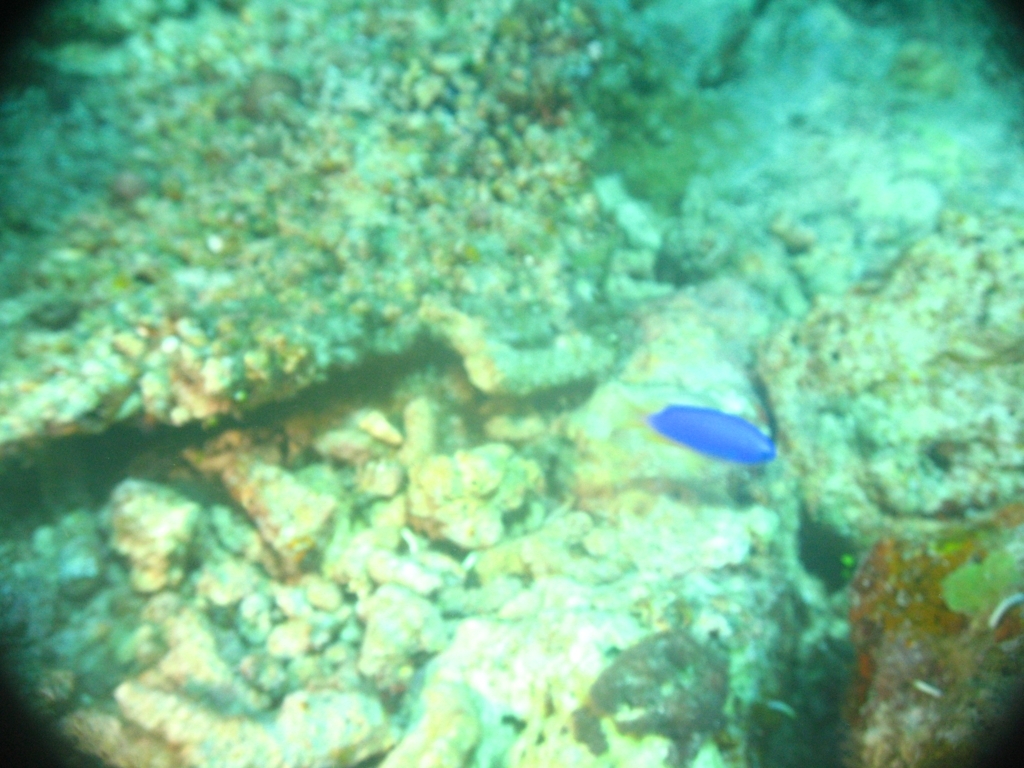}     &   &   \includegraphics[width=0.19\textwidth]{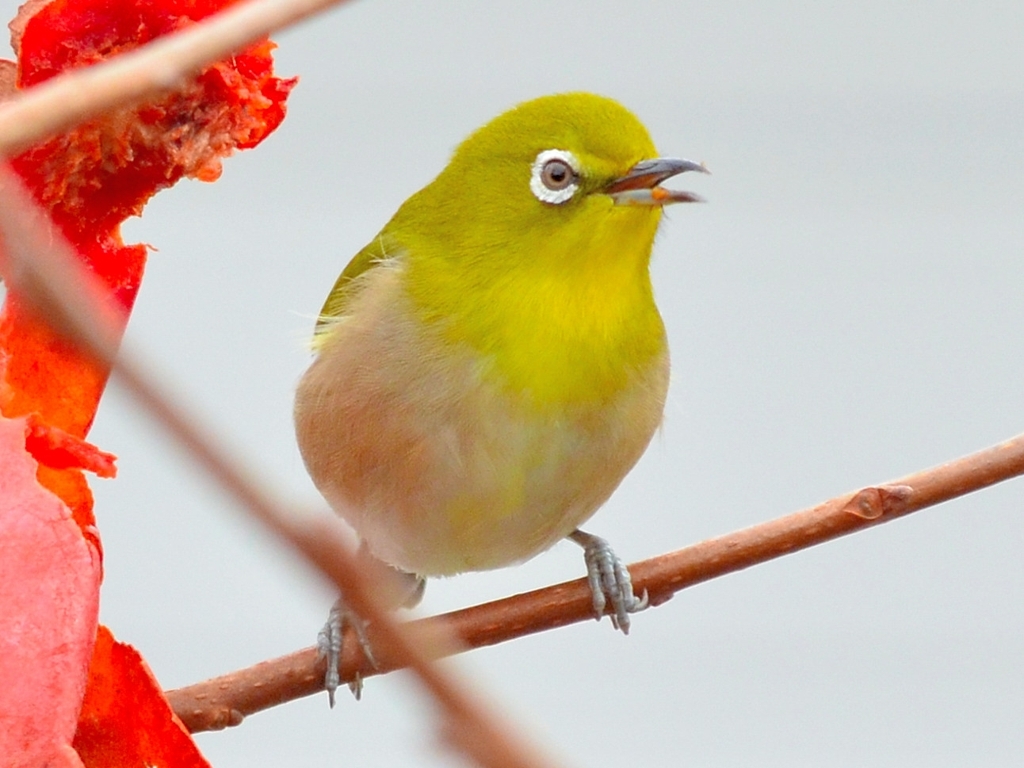}     &  \includegraphics[width=0.19\textwidth]{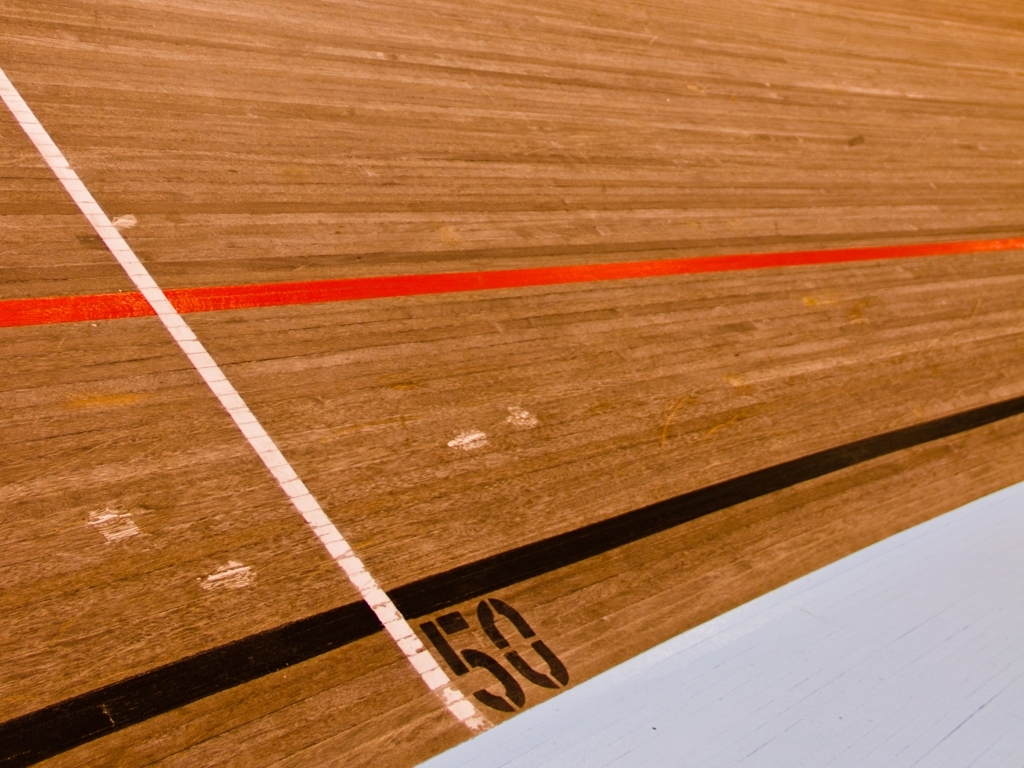} \\
(a) & (b) & & (c) & (d) \\ 
{\textbf{28.97}} & {\textbf{24.49}} & {\color[HTML]{34696D} \textbf{\begin{tabular}[c]{@{}c@{}}Human\\ Opinion Score\end{tabular}}} & { \textbf{75.63}} & { \textbf{65.15}} \\ \hdashline
{\color[HTML]{FE0000} \textbf{40.56}} & {\color[HTML]{009901} \textbf{25.43}} & {\color[HTML]{999903} \textbf{\begin{tabular}[c]{@{}c@{}}$\text{GRepQ}_\text{D}$ (LL)\\ Prediction\end{tabular}}}  & {\color[HTML]{FE0000} \textbf{62.04}} & {\color[HTML]{009901} \textbf{61.29}} \\ \hdashline
{\color[HTML]{009901} \textbf{27.66}} & {\color[HTML]{FE0000} \textbf{39.40}} & {\color[HTML]{986536} \textbf{\begin{tabular}[c]{@{}c@{}}$\text{GRepQ}_\text{D}$ (HL)\\ Prediction\end{tabular}}} & {\color[HTML]{009901} \textbf{72.33}} & {\color[HTML]{FE0000} \textbf{49.85}}
\end{tabular}
\captionof{figure}{Demonstrating the complementarity of high and low-level model predictions. Images with different MOS from the KonIQ-10K database are listed. Low and high-level model predictions are mentioned below their respective MOS. Predictions that agree with human opinions are marked with {\color[HTML]{009901}green} while erroneous predictions are marked with \textcolor{red}{red}.}
\label{fig:comp_images}
\end{table*}

\noindent \textbf{Analyzing High-Level Feature Representations:} We analyze the impact of our group-contrastive learning in improving the high-level quality representations. For this analysis, we identify extremely good and extremely bad quality images based on mean opinion score (MOS) greater than 75 or less than 25 respectively on the combined CLIVE and KonIQ datasets. We show the feature representations of the CLIP model in \cref{fig:tsne}a and those of our model in \cref{fig:tsne}b. We see that our learned representations are better separable between the higher and lower-quality images. This leads to the superior performance of our high-level model when compared to CLIP-IQA. \\

\begin{figure}[!t]
    \centering
    \includegraphics[trim={0cm 0cm 0cm 2.5cm}, clip, scale=0.21]{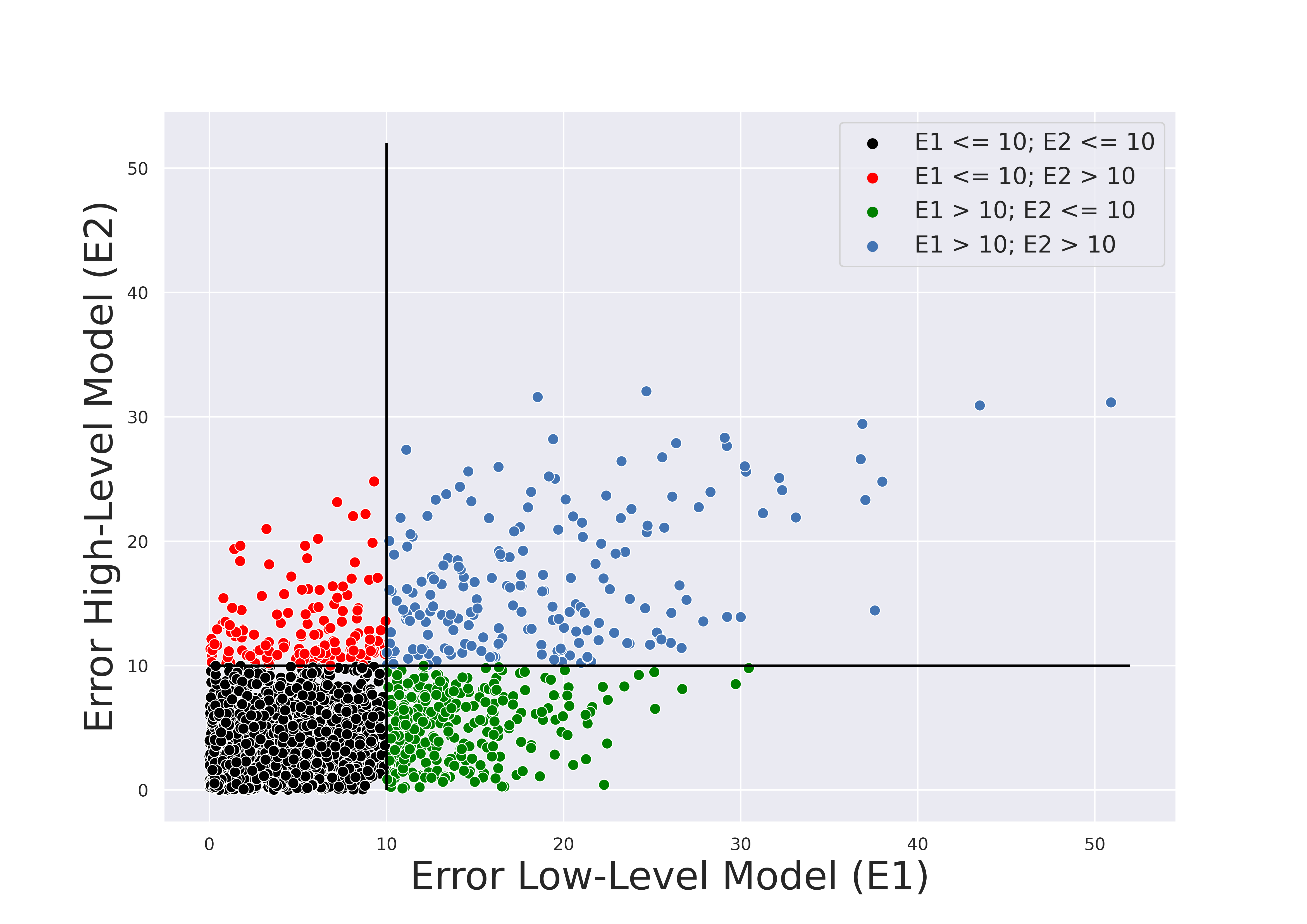}  
    \caption{Analysis of absolute error in quality predictions from high and low-level models on the KonIQ-10K database.}
    \label{fig:error}
\end{figure}

\noindent \textbf{Complementarity of High and Low-Level Features:} We present a qualitative and quantitative analysis of the complementarity of representations from both encoders. 
We show examples of when the two models outperform each other in  \cref{fig:comp_images}. For instance, \cref{fig:comp_images}a shows that the low-level model makes an erroneous prediction since only the object in focus is blurred, but the background is relatively clean. \cref{fig:comp_images}d shows that the image does not contain enough contextual information for the high-level model to make an accurate prediction. We also perform an error-based feature complementarity analysis in \cref{fig:error}. In particular, we compute the absolute error between the MOS predicted by the high and low-level models and the true MOS and show them in four quadrants. We see several examples where one of the models performs much better than the other. This shows that the models have complementary behavior in many examples. \\

\noindent \textbf{Limitations:} In the low-data setting, the low-level model does not perform as well as the high-level model on in-the-wild datasets. Since the low-level model is more suited to capture varied distortion levels rather than content, synthetic datasets benefit more from this model. Secondly, the high-level model uses fixed prompts and can be further improved through prompt engineering or tuning. 

\section{Concluding Remarks}\label{sec:conclusion}

We design generalizable low-level and high-level quality representations that enable IQA in a data-efficient setting. Specifically, we learn low-level features using a novel quality-aware contrastive learning strategy that is distortion-agnostic. Secondly, we present a group-contrastive learning framework that learns to elicit semantic-based high-level quality information from images. We show that both sets of representations lead to accurate prediction of quality scores in both the data-efficient and zero-shot settings on diverse datasets. This demonstrates the generalizability of our learned features. Future advances in self-supervised learning and quality-specific prompt engineering could be used to further enhance the generalizability of models for data-efficient NR IQA. \\

\noindent\textbf{Acknowledgement:} This work was supported in part by Department of Science and Technology, Government of India under grant CRG/2020/003516.

{\small
\bibliographystyle{ieee_fullname}
\bibliography{main}
}

\end{document}